%% file: sample-sigconf.tex
\documentclass[sigconf]{acmart}
\input{tikz_package}

\usepackage{tikz}
\usepackage{color}
\usepackage{tikz-dependency}
\usetikzlibrary{shapes, arrows, positioning, decorations.markings}
\definecolor {processblue}{cmyk}{0.96,0,0,0}
\tikzstyle{int}=[draw, fill=blue!20, minimum size=2em]
\tikzstyle{init} = [pin edge={to-,thin,black}]
\usetikzlibrary{shapes}
\usetikzlibrary{fit}
\usetikzlibrary{chains}
\usetikzlibrary{arrows}
\tikzset{
semi/.style={
  semicircle, ,top color =white , bottom color = processblue!20 ,
draw, processblue , text=blue,
  draw,
  minimum size=0.3cm
  }
}
\tikzstyle{plate} = [draw, rectangle, rounded corners, fit=#1]
\tikzstyle{wrap} = [inner sep=0pt, fit=#1]
\tikzstyle{caption} = [node distance=0] %
\tikzstyle{bottom plate caption} = [caption, node distance=0, inner sep=0pt,
below left=-5pt and 0pt of #1.south east] %
\tikzstyle{top plate caption} = [caption, node distance=0, inner sep=0pt,
below left=0pt and 0pt of #1.north east] %
\usepackage{tikz}
\usetikzlibrary{shapes,arrows,shadows}
\usepackage{amsmath,bm,times}
\AtBeginDocument{%
  \providecommand\BibTeX{{%
    \normalfont B\kern-0.5em{\scshape i\kern-0.25em b}\kern-0.8em\TeX}}}

\copyrightyear{2023}
\acmYear{2023}
\setcopyright{acmlicensed}\acmConference[WWW '23]{Proceedings of the ACM Web Conference 2023}{May 1--5, 2023}{Austin, TX, USA}
\acmBooktitle{Proceedings of the ACM Web Conference 2023 (WWW '23), May 1--5, 2023, Austin, TX, USA}
\acmPrice{15.00}
\acmDOI{10.1145/3543507.3583325}
\acmISBN{978-1-4503-9416-1/23/04}

\begin{document}

\title{Self-training through Classifier Disagreement for Cross-Domain Opinion
Target Extraction}


\author{Kai Sun}
\authornote{Both authors contributed equally to the paper}
\email{sunkai@act.buaa.edu.cn}
\affiliation{%
  \institution{ Beijing Advanced Innovation Center for Big Data and Brain Computing, School of Computer Science and Engineering, Beihang University}
  \city{Beijing}
  \country{China}
}

\author{Richong Zhang}
\authornote{Corresponding Author}
\email{zhangrc@act.buaa.edu.cn}
\affiliation{%
  \institution{ Beijing Advanced Innovation Center for Big Data and Brain Computing, School of Computer Science and Engineering, Beihang University}
  \city{Beijing}
  \country{China}
}

\author{Samuel Mensah}
\authornotemark[1]
\email{s.mensah@sheffield.ac.uk}
\affiliation{%
  \institution{ Department of Computer Science,\\ University of Sheffield,}
   \city{Sheffield}
  \country{United Kingdom}
}

\author{Nikolaos Aletras}
\email{n.aletras@sheffield.ac.uk}
\affiliation{%
  \institution{ Department of Computer Science,\\ University of Sheffield,}
  \city{Sheffield}
  \country{United Kingdom}
}

\author{Yongyi Mao}
\email{yymao@site.uottawa.ca}
\affiliation{%
 \institution{ School of Electrical Engineering and Computer Science, University of Ottawa}
   \city{Ottawa}
 \country{Canada}
}

\author{Xudong Liu}
\email{liuxd@act.buaa.edu.cn}
\affiliation{%
  \institution{ Beijing Advanced Innovation Center for Big Data and Brain Computing, School of Computer Science and Engineering, Beihang University}
  \city{Beijing}
  \country{China}
}

\renewcommand{\shortauthors}{Sun et al.}

\begin{abstract}
 Opinion target extraction (OTE) or aspect extraction (AE) is a fundamental task in opinion mining that aims to extract the targets (or aspects) on which opinions have been expressed. Recent work focus on cross-domain OTE, which is typically encountered in real-world scenarios, where the testing and training distributions differ. Most methods use domain adversarial neural networks that aim to reduce the domain gap between the labelled source and unlabelled target domains to improve target domain performance. However, this approach only aligns feature distributions and does not account for class-wise feature alignment, leading to suboptimal results. Semi-supervised learning (SSL) has been explored as a solution, but is limited by the quality of pseudo-labels generated by the model. Inspired by the theoretical foundations in domain adaptation~\cite{ben2010theory}, we propose a new SSL approach that opts for selecting target samples whose model output from a domain-specific teacher and student network disagree on the unlabelled target data, in an effort to boost the target domain performance. Extensive experiments on benchmark cross-domain OTE datasets show that this approach is effective and performs consistently well in settings with large domain shifts.
\end{abstract}



\received{20 February 2007}
\received[revised]{12 March 2009}
\received[accepted]{5 June 2009}

\maketitle

\section{Introduction}\label{introduction}
% \textcolor{red}{(Lets add some background information to relate to web content. like something below.... i copied them from a journal's first paragraph. Lets stress the importance of studying the cross domain OTE task.)}

% \textcolor{red}{(The internet has recently experienced an exceptionally rapid growth of user generated content, and this is partly due to the development of online applications including e-commerce and social media. Reviews, in particular, generated by users often contain information relating to sentiments expressed on different aspects of products. This information is of utmost benefit to businesses that aim to improve their products. By analyzing reviews, merchants or manufacturers can obtain public opinions on products and services in order to improve their competitiveness or help to make the best business decision. On the other hand, consumers can also decide whether a product is suitable for purchase based on the reviews provided by purchasers. This scenario has led to the development of sentiment analysis \cite{10.1145/3529954} and its application in recommender systems \cite{10.1145/3473970}.)}

The growth of e-commerce websites has allowed consumers to directly interact with products, leading to an increase in user-generated content. In particular, reviews of products generated by users has grown at an astronomical rate with the increasingly accessibility and affordability of the internet. These reviews, which are often expressed as text, contain sentiment information or opinion words expressed on different aspects of products (referred to as opinion targets or aspect terms). As a result, opinion words along with its corresponding aspect terms have become an important resource to improve recommender systems for Web resource discovery. A typical example is the recommendation of a book in Goodreads \cite{Goodreads} based on the opinions expressed on a specific section in the book.

% \textcolor{red}{(Sentiment analysis, also known as opinion mining, aims to identify the sentiment polarity (e.g., positive, negative or neutral) of a piece of text. It is a fundamental task in natural language processing (NLP), and has been used extensively for commodity online review analysis, public opinion analysis, etc. In this research field,)}

% Opinion target extraction (OTE) is a fundamental task in sentiment analysis which has great significance in various applications ranging from opinion mining \cite{pang2008opinion} to aspect-level sentiment analysis \cite{schouten2015survey}.

This phenomenon has led to the increased research in opinion mining~\cite{DBLP:books/daglib/0036864}. This paper focuses on opinion target extraction (OTE) (or aspect extraction (AE)), a fundamental step in opinion mining. AE aims to extract from opinionated sentences the aspects on which opinions have been expressed. Traditional approaches \cite{DBLP:conf/emnlp/JakobG10,DBLP:conf/naacl/YangE15} utilize hand-crafted features, which heavily rely on feature extraction. With the advances in deep learning, recent approaches \cite{DBLP:conf/acl/MaLWXW19,DBLP:conf/acl/LiCQLS20,DBLP:conf/acl/WeiHZCY20} are based on neural networks that are trained in a supervised manner. However, as with any other supervised learning method, these approaches perform poorly when there is a change in domain upon deployment.  Cross-domain OTE \cite{DBLP:conf/aaai/DingYJ17} has emerged as a solution by using unsupervised domain adaptation (UDA) techniques~\cite{DBLP:journals/tkde/BollegalaWC13,DBLP:conf/acl/ZhouXHH16,DBLP:conf/emnlp/HeLND18}  to reduce the domain shift between a labelled source and unlabelled target domain.

%Furthermore, the idea of collecting labelled target data for domain shifts is prohibitively expensive for the learning task.

One typical line of work aims to reduce domain shifts via domain adversarial neural networks (DANN)~\cite{DBLP:journals/jmlr/GaninUAGLLML16}. Given a labelled source and unlabelled target domain data, DANNs attempt to learn representations that are discriminative on the source domain and invariant to the domain distribution. However, DANNs align the feature distributions of the source and target data inputs (i.e., aligning the marginal distribution), neglecting the feature alignment at the class-level \cite{tan2019generalized}. As a consequence, the resulting target features are non-discriminative with respect to the class labels, which consequently leads to suboptimal target domain performance.

Semi-supervised learning (SSL)~\cite{chapelle2009semi} has been explored to learn target discriminative features by generating pseudo-labels from the unlabelled target data. While SSL approaches have been heavily employed to boost domain adaptation in vision tasks~\cite{tarvainen2017mean,ke2019dual,xie2020self}, it has been lightly touched in cross-domain OTE~\cite{DBLP:conf/acl/YuGX21,zhou2021adaptive}. The state-of-the-art method Adaptive Hybrid Framework~(AHF) \citep{zhou2021adaptive} adapts a mean teacher (i.e., teacher and student networks) \cite{tarvainen2017mean} into the task. The teacher is modelled as a feedforward network while the student is a DANN (i.e., developed by augmenting the feedforward network with a discriminator). Here, knowledge on the target's output of the teacher-student networks is shared among the networks to learn the target-discriminative features.  Although AHF demonstrates the importance of SSL, the fundamental weakness of the mean-teacher cannot be ignored. Specifically, \citet{ke2019dual} provided theoretical and empirical proof to show that the weights of the teacher quickly converges to that of the student as training progresses, which consequently leads to a performance bottleneck.

\begin{figure}[ht]
	\centering
	\renewcommand\tabcolsep{0pt}
	\begin{tabular}{c}
	\scalebox{0.65}{\includegraphics[]{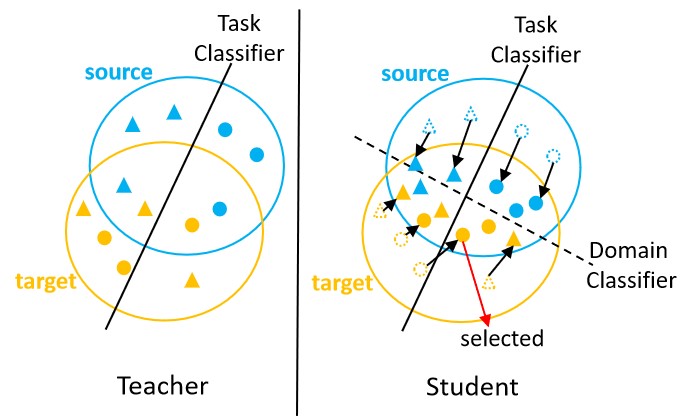}}
	\end{tabular}
	\caption{Illustrative example of source and target distributions induced by a teacher and student network (Best viewed in color). Target samples that change class due to adversarial learning by the student network are selected to self-train the student.}
	\label{fig:motivation}
\end{figure}

%SCD achieves this by comparing the two target distributions induced separately by the student and teacher networks. Fig.~\ref{fig:motivation} illustrates the two target distributions' difference. These high-quality pseudo-labelled target samples are those that disagree (i.e., discrepancy in target predictions) with their correspondence in the teacher feature space.

These findings motivate us to decouple the student-teacher networks and optimize the networks through independent paths to prevent the networks from collapsing into each other \cite{ke2019dual}. We propose a novel SSL approach, which performs Self-training through Classifier Disagreement (SCD), to effectively explore the outputs of the student and teacher networks on the unlabelled target domain. SCD is inspired by the theory of domain adaptation~\cite{ben2010theory}, which allows us to detect high-quality pseudo-labelled target samples in the student feature space to self-train the student for cross-domain OTE. As demonstrated in Fig.~\ref{fig:motivation}, SCD achieves this by comparing the two target distributions induced separately by the student and teacher networks. These high-quality pseudo-labelled target samples are those that disagree (i.e., discrepancy in target predictions) with their correspondence in the teacher feature space. We perform extensive experiments and find that SCD not only achieves impressive performance but also performs consistently well in large domain shifts on cross-domain OTE.

Our contribution can be summarized as follows:

\begin{itemize}
    \item We develop a novel SSL approach for cross-domain OTE, referred to as Self-training through Classifier Disagreement (SCD) which leverages high-quality pseudo-labelled target samples in the student feature space to improve target performance in cross-domain OTE.
    \item We demonstrate that SCD is favourable in large domain divergence - a key direction in the domain adaptation research.
    \item We perform extensive experiments and show that SCD achieves state-of-the-art results in nine out of ten transfer pairs for the cross-domain OTE task.
 
\end{itemize}

\section{Related Work}\label{relworks}

There is a growing literature on OTE~\cite{DBLP:conf/ijcai/LiBLLY18,DBLP:conf/acl/XuLSY18,DBLP:conf/aaai/LiBLL19,DBLP:conf/acl/MaLWXW19,DBLP:conf/acl/LiCQLS20,DBLP:conf/acl/WeiHZCY20} but they mostly focus on single domain learning. However, in real-world scenarios, the training distribution used by a classifier may differ from the test distribution, which is a big challenge for single domain learning methods.

% \textcolor{red}{Traditional methods hand-crafted domain-independent features, and used Conditional Random Fields (CRFs)~\cite{DBLP:conf/emnlp/JakobG10,DBLP:conf/acl/LiPJYZ12,DBLP:conf/semeval/Chernyshevich14}. While hand-crafted features are useful, they are manually engineered and require human experts, making them time-consuming and expensive to obtain.}

% \textcolor{red}{\cite{DBLP:conf/aaai/DingYJ17} use domain-independent rules to generate auxiliary labels and use a recurrent neural network to learn a domain-invariant hidden representation for each word. However, the manually defined rules have limited coverage.  \cite{DBLP:conf/acl/PanW18} introduce an opinion word extraction as an auxiliary task based on a critical assumption that associative patterns exist between opinion targets and opinion words irrespective of the domain. They use syntactic relations in dependency trees as the pivot to bridge the domain gap for cross-domain OTE. One downside is that these external linguistic resources are derived from traditional NLP systems which may propagate errors. }

%\textcolor{red}{These} Cross-domain OTE models mitigate the issue of acquiring  manually labeled data that can be expensive to obtain. They also extend single-domain OTE models \cite{DBLP:conf/ijcai/LiBLLY18,DBLP:conf/acl/MaLWXW19} that may suffer from domain dependency.

Cross-domain learning has been explored for the OTE task. Traditional methods use hand-crafted domain-independent features and use Conditional Random Fields (CRFs)~\cite{DBLP:conf/emnlp/JakobG10,DBLP:conf/acl/LiPJYZ12,DBLP:conf/semeval/Chernyshevich14}. While hand-crafted features are useful, they are manually engineered and require human experts, making them time-consuming and expensive to obtain. So far, some neural models have been proposed for cross-domain OTE~\cite{DBLP:conf/aaai/DingYJ17,DBLP:conf/acl/PanW18,DBLP:conf/emnlp/LiLWBZY19,DBLP:conf/emnlp/GongYX20,DBLP:conf/acl/0002Q20,DBLP:conf/acl/YuGX21,zhou2021adaptive}. The common paradigm in prior work is to reduce the domain shift between the source and target domains. Among recent work, \citet{DBLP:conf/aaai/DingYJ17} proposed a hierarchical network trained with joint
training (Hier-Joint). This method uses  domain-independent rules to generate auxiliary labels and use a recurrent neural network to learn a domain-invariant hidden representation for each word. However, the manually defined rules have limited coverage. A similar method, namely, Recursive Neural Structural Correspondence Network (RNSCN)~\cite{DBLP:conf/acl/PanW18} introduces an opinion word extraction as an auxiliary task based on a critical assumption that associative patterns exist between aspect terms and opinion words irrespective of the domain. They use syntactic relations in dependency trees as the pivot to bridge the domain gap for cross-domain OTE. However, the external linguistic resources used are derived from traditional feature-engineered NLP systems which may propagate errors. More recent methods, including the Aspect Boundary Selective Adversarial Learning model (AD-SAL)~\cite{DBLP:conf/emnlp/LiLWBZY19} uses an adversarial network with attention mechanisms  to learn domain-invariant features. \citet{DBLP:conf/emnlp/GongYX20} proposed BERT$_{\rm E}$-UDA to integrate BERT fine-tuned on domain information for the task. \citet{DBLP:conf/acl/0002Q20} proposed a Semantic Bridge network (SemBridge) which constructs syntactic and semantic bridges to transfer common knowledge across domains.

While significant progress has been made, the majority of the proposed models neglect the feature distribution alignment at the class-level. Hence, their performance cannot be guaranteed because they do not learn target discriminative features. Recently, a Cross-Domain Review Generation model based on BERT (BERT$_{\rm E}$-CDRG)~\cite{DBLP:conf/acl/YuGX21} generated target domain data with fine-grained annotations aiming to learn the target discriminative features. Perhaps, AHF~\cite{zhou2021adaptive} is the first to use SSL in the task. AHF adapts a mean teacher in which the teacher and student networks are found to be tightly coupled during training, leading to a performance bottleneck~\cite{ke2019dual}. Elsewhere, researchers have delicately designed SSL approaches that allow individual models to iteratively learn from each other, thus, preventing these models from collapsing into each other \cite{qiao2018deep,ke2019dual,chen2021semi}. Such approaches have demonstrated substantial improvements over the mean-teacher.

\section{Preliminaries}

Our method is inspired by the theory of domain adaptation proposed by \citet{ben2010theory}, which provides an upper bound on the target error in terms of the source error and the domain divergence. Suppose $h\in\mathcal{H}$ is a hypothesis, \citet{ben2010theory} theorized that the target error $\epsilon_{\mathcal{T}}(h)$ (which can also be viewed as the target performance) is bounded by the source error $\epsilon_{\mathcal{S}}(h)$ (i.e., the source performance) and the symmetric difference hypothesis divergence $\mathcal{H}\Delta\mathcal{H}$-divergence  between the source $\mathcal{S}$ and target $\mathcal{T}$ distributions, denoted as $d_{\mathcal{H}\Delta\mathcal{H}}(\mathcal{S},\mathcal{T})$ (i.e., a measure of the domain shift). Formally,

\begin{equation}
\begin{aligned}
\forall h\in \mathcal{H}, \epsilon_{\mathcal{T}}(h) \leq \epsilon_{\mathcal{S}}(h) + \frac{1}{2}d_{\mathcal{H}\Delta\mathcal{H}}(\mathcal{S},\mathcal{T}) + \beta
\end{aligned}
\label{eqn:error-bound}
\end{equation}
 
\noindent where $\beta$ is the optimal joint error on the source and target domains which should be small for domain adaptation. Note, $\beta$ is a constant which is independent of $h$. To obtain a better estimate of $\epsilon_{\mathcal{T}}(h)$, a learner can either reduce the source error $\epsilon_{\mathcal{S}}(h)$ or/and the divergence $d_{\mathcal{H}\Delta\mathcal{H}}(\mathcal{S},\mathcal{T})$, which can be estimated from finite samples of the source and target domains \cite{ben2010theory}. 

\section{Problem Statement}\label{problem}
 
The OTE task is formulated as a sequence labeling problem. Given the $j$-th input sentence $\mathbf{x}_j=\{x_{ij}\}_{i=1}^{n}$ with $n$ words, the word $x_{ij}$ is represented as a feature vector. The goal is to predict the label sequence $\mathbf{y}_j=\{y_{ij}\}_{i=1}^{n}$, with $y_{ij}\in \mathcal{Y}=\{\rm B, I, O\}$, denoting the {\bf B}eginning, {\bf I}nside and {\bf O}utside of an opinion target or aspect term.

In this paper, we focus on the cross-domain setting which is typically tackled through unsupervised domain adaptation (UDA). Particularly, UDA aims to transfer knowledge from a labelled source domain to an unlabelled target domain, whose data distribution has a considerable shift from that of the source domain. Formally, suppose a labelled source domain dataset with $N_\mathcal{S}$ sentence and label pairs $D_\mathcal{S} = \{(\mathbf{x}_j^\mathcal{S},\mathbf{y}_j^\mathcal{S})\}_{j=1}^{N_\mathcal{S}}$, and  an unlabeled dataset in a target domain with $N_\mathcal{T}$ unlabelled sentences $D_{\mathcal{T}} =  \{(\mathbf{x}_j^\mathcal{T}) \}_{j=1}^{N_\mathcal{T}}$. Our goal is to predict labels of testing samples in the target domain using a model trained on $D_\mathcal{S} \cup D_\mathcal{T}$. \footnote{Hereinafter, subscripts or superscripts are omitted for clarity, and the term ``aspect'' will be used instead of ``opinion target'' to avoid confusion with the target domain.}

% \footnote{In our model description, if no confusion arises, superscripts or subscripts are dropped for clarity and brevity. }

\section{Methodology}\label{model}

Our method is based on a teacher-student network structure. Teacher $A$ learns on the source data $D_{\mathcal{S}}$; and Student $B$  learns on both the source $D_{\mathcal{S}}$ and target domain data $D_{\mathcal{T}}$. Both trained networks generate pseudo-labelled target samples on the unlabelled target domain, which are then compared to detect high quality pseudo-labelled target samples to self-train the student for cross-domain OTE.

\subsection{Teacher Network}

The teacher network $A=\{A_e,A_l\}$ is a neural network, consisting of a feature encoder $A_e$ and a label classifier $A_l$. In our work, $A_e$ is modelled using a BiLSTM~\cite{hochreiter1997long} or BERT~\cite{Devlin2019BERTPO} since they are both widely used approaches for sequence labelling problems. $A_l$ on the other hand is modelled using a softmax function. Although the CRF~\cite{lafferty2001conditional} is a typical choice to model the label classifier for sequence labelling problems, the softmax offers comparable performance in cross-domain OTE~\cite{DBLP:conf/emnlp/LiLWBZY19}. Hence, given the sentence $\mathbf{x}_j=\{x_{ij}\}_{i=1}^{n}$, $A_e$ extracts the context features $\mathbf{f}^{A_e}_j=\{f^{ A_e}_{ij}\}_{i=1}^n$. Now, for each word-level feature $f^{A_e}_{ij}$, the label classifier $A_l$ is  applied to output the prediction probability $P(\hat{y}^{A_l}_{ij})$ over the tag set $\mathcal{Y}$. As the teacher is trained over the source data only, the classification loss by the teacher network is given by:

\begin{equation}
\mathcal{L}^{A}_y = \frac{1}{N_\mathcal{S}}\sum_{j=1}^{N_\mathcal{S}}\sum_{i=1}^{n}\ell(P(\hat{y}^{A_l}_{ij}), y_{ij})
\label{eqn:source-only}
\end{equation}

where $P(\hat{y}_{ij}^{A_l})$ is the probability prediction for the word $x_{ij}\in \mathbf{x}_j^{\mathcal{S}}$ and $y_{ij}\in \mathbf{y}_j^{\mathcal{S}}$ is the ground-truth of $x_{ij}$. $\ell$ is the cross-entropy loss function.

 Now suppose $\mathbf{F}^{{A}_e}_\mathcal{S}$ and  $\mathbf{F}^{{A}_e}_\mathcal{T}$ are fixed representations of the respective source and target domain data produced by the trained teacher $A_e$. The upper bound on the target error $\epsilon_{\mathcal{T}}({A}_l)$ of the label classifier $A_l$ can be expressed as:

\begin{equation}
\epsilon_{\mathcal{T}}({A}_l) \leq \epsilon_{\mathcal{S}}({A}_l) + \frac{1}{2}d_{\mathcal{H}\Delta\mathcal{H}}(\mathbf{F}^{{A}_e}_\mathcal{S},\mathbf{F}^{{A}_e}_\mathcal{T}) + \beta  
\label{eqn:error-bounds-Cy-src}
\end{equation}

It is easy to see that the teacher network simply reduces the source error  $\epsilon_{\mathcal{S}}({A}_l)$ by \eqref{eqn:source-only} while the domain shift $d_{\mathcal{H}\Delta\mathcal{H}}(\mathbf{F}^{{A}_e}_\mathcal{S},\mathbf{F}^{{A}_e}_\mathcal{T})$ remains large since the network does not have an appropriate component  to reduce the domain shift. This leads to a suboptimal estimate for the bound of the target errors $\epsilon_{\mathcal{T}}({A}_l)$. 
 
\subsection{Student Network}

As we have seen in the previous section, the teacher applies domain-specific knowledge (i.e., the source domain) for inference, which may underperform on the target domain due to difference in the data distribution. Ideally, the network should have the ability to perform in different domains. We introduce the student network as a solution.

The student network is analogous to a student who learns several subjects simultaneously in order to perform well in those subjects. This is different from teachers who are normally experts in a single subject. This implies that the student network not only desires to be as excellent as the domain-specific teacher on the source data but also aims to perform well on the target data. To this end, the student network is developed by augmenting a teacher network with a discriminator (or domain classifier), following DANN \cite{DBLP:journals/jmlr/GaninUAGLLML16}. Accordingly, the student network $B=\{B_e,B_l,B_d\}$ consists of a feature encoder $B_e$; label classifier $B_l$; and domain classifier $B_d$, which determines if the sample comes from the source or target domain. $B_e$ extracts the context features $\mathbf{f}^{B_e}_j$ from the sentence $\mathbf{x}_j\in D_\mathcal{S}\cup D_\mathcal{T}$ and feeds to $B_l$ to learn discriminative features on the source domain, following a similar classification loss with Eqn.~\eqref{eqn:source-only}. Formally, the classification loss is defined as:

\begin{equation}
\mathcal{L}^{B}_y = \frac{1}{N_\mathcal{S}}\sum_{j=1}^{N_\mathcal{S}}\sum_{i=1}^{n}\ell(P(\hat{y}^{B_l}_{ij}), y_{ij})
\label{eqn:adv-only}
\end{equation}

where $P(\hat{y}^{B_l}_{ij})$ is the probability prediction for the word $x_{ij}\in\mathbf{x}_j^{\mathcal{S}}$ and $y_{ij}\in \mathbf{y}_j^{\mathcal{S}}$ is the ground-truth. At the same time, $\mathbf{f}^{B_e}_j$ is fed to a domain classifier $B_d$ to learn domain-invariant features through a gradient reversal layer (GRL) \cite{DBLP:journals/jmlr/GaninUAGLLML16}. Formally, the GRL $R_{\lambda}(\cdot)$ acts as an identity function in the forward pass, i.e., $R_{\lambda}(\mathbf{f}^{B_e}_j)=\mathbf{f}^{B_e}_j$,  and backpropagates the negation of the gradient in the backward pass, i.e., $\partial R_{\lambda}(\mathbf{f}^{B_e}_j) / \partial  \mathbf{f}^{B_e}_j=-\lambda I$. Consequently, $B_e$ maximizes the domain classification loss $\mathcal{L}_d^B$ through the GRL while $B_d$ minimizes $\mathcal{L}_d^B$ to make $\mathbf{f}^{B_e}_j$ domain-invariant. The domain classification loss $\mathcal{L}_d^B$ is defined as follows:

{\small
\begin{align}
 \begin{split}
\mathcal{L}_{d}^B=\sum_{j=1}^{N}d_j{\rm log}(P(\hat{d}^{B_d}_j))+(1-d_j){\rm log}(1-P(\hat{d}^{B_d}_j))
 \end{split}
\label{eqn:loss_adv}
\end{align}
}

\noindent where $d_j=1$ indicates that the $j$-th sentence comes from the source domain, otherwise $d_j=0$; $P(\hat{d}^{B_d}_j)$ is the domain probability prediction of the sentence-level feature $\mathbf{x}_j$; $N=N_\mathcal{S}+N_\mathcal{T}$.

Suppose $\mathbf{F}^{{B}_e}_\mathcal{S}$ and  $\mathbf{F}^{{B}_e}_\mathcal{T}$ are fixed representations of the respective source and target domain data produced by the trained student encoder $B_e$. The upper bound on the student label classifier $B_l$ can be expressed as:

\begin{equation}
\epsilon_{\mathcal{T}}({B}_l) \leq \epsilon_{\mathcal{S}}({B}_l) + \frac{1}{2}d_{\mathcal{H}\Delta\mathcal{H}}(\mathbf{F}^{{B}_e}_\mathcal{S},\mathbf{F}^{{B}_e}_\mathcal{T}) + \beta  
\label{eqn:error-bounds-Cy}
\end{equation}

The source error $\epsilon_{\mathcal{S}}({B}_l)$ is comparable with $\epsilon_{\mathcal{S}}({A}_l)$ since the student and teacher are trained on the source data using the same network pipeline (comparing \eqref{eqn:source-only} and \eqref{eqn:adv-only}, and also empirically demonstrated in Table~\ref{tab:source-results}).   But the student network has been shown to reduce the domain divergence with a theoretical guarantee via the GRL~\cite{DBLP:journals/jmlr/GaninUAGLLML16}. This means $d_{\mathcal{H}\Delta\mathcal{H}}(\mathbf{F}^{{B}_e}_\mathcal{S},\mathbf{F}^{{B}_e}_\mathcal{T})$ is relatively small, i.e., $d_{\mathcal{H}\Delta\mathcal{H}}(\mathbf{F}^{{B}_e}_\mathcal{S},\mathbf{F}^{{B}_e}_\mathcal{T}) \leq d_{\mathcal{H}\Delta\mathcal{H}}(\mathbf{F}^{{A}_e}_\mathcal{S},\mathbf{F}^{{A}_e}_\mathcal{T})$, and therefore leads to a better estimate of $\epsilon_{\mathcal{T}}({B}_l)$. In other words, the student performs better than the domain-specific teacher on the target data due to the mitigation of the domain shift.

\subsection{Self-training through Classifier Disagreement}

\begin{figure}[htbp]
\centering
	\renewcommand\tabcolsep{0pt}
	\begin{tabular}{c}
	\scalebox{0.68}{\input{model_pseudo.tex}}\\
	\end{tabular}
	\caption{Overview of our SSL Approach. Both Teacher and Student networks have been earlier trained by Eqn. \eqref{eqn:source-only}, \eqref{eqn:adv-only} and \eqref{eqn:loss_adv}. The Student network alone is further self-trained through classifier disagreement on the target domain. This figure is best viewed in color.}
	\label{fig:models}
\end{figure}
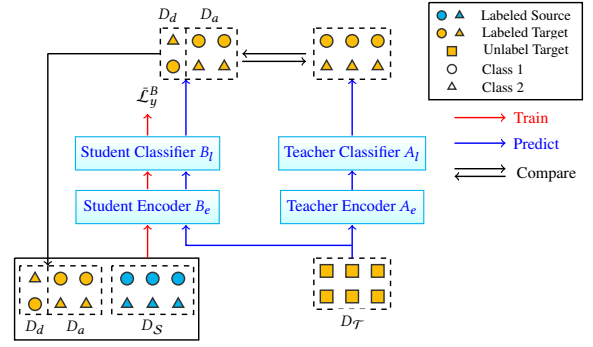

The student network improves target performance by aligning the source and target data distributions. It just so happens that it simply aligns the data distribution without considering the alignment at the class-level~\cite{tan2019generalized}, leading to suboptimal performance. Such a situation occurs due to the lack of labelled target data to learn target discriminative features. The fundamental challenge is that we do not have access to labelled target data.

To this end, we introduce a strikingly simple approach to collect high-quality pseudo-labelled target samples  to improve the class-level alignment of the student network. Fig \ref{fig:models} shows an overview of our approach, which we refer to as Self-training through Classifier Disagreement (SCD). Suppose the trained student and teacher networks (i.e., trained by Eqn. \eqref{eqn:source-only}, \eqref{eqn:adv-only} and \eqref{eqn:loss_adv}) assign pseudo-labels to the unlabelled target data. Eqns \eqref{eqn:error-bounds-Cy-src} and \eqref{eqn:error-bounds-Cy} indicate that the increase in target performance by the student can be explained by the target samples that have shifted toward the domain-invariant feature space (i.e., the student feature space). Our goal is to self-train the student network by leveraging the target samples responsible for the performance improvement in the target domain.

This strategy is only beneficial if the domain shift is large since this will lead to a large set of high-quality pseudo-labelled target samples. Otherwise, both networks will have comparable performance on the unlabelled target domain and the performance gain is minimal. To extend the approach to problems with close similarity between domains, we split the self-training learning problem by paying attention to: 1) $D_d$, the target samples in the student feature space that {\it disagree} with their counterpart in the teacher feature space; and 2) $D_a$, the target samples in the student feature space that {\it agree} with their counterpart in the teacher feature space.

Formally, let us suppose the student and teacher networks are already trained (i.e., without self-training). As we aim to self-train the Student network, we can rewrite the classification loss expressed in \eqref{eqn:adv-only} as $\mathcal{L}^{B(0)}_y$ to represent the initial classification loss of the Student network.  Now, let us suppose the teacher and student networks assign the pseudo-labels $\bar{\mathbf{y}}_j^{A_l}=\{\bar{y}_{ij}^{A_l}\}_{i=1}^n$ and $\bar{\mathbf{y}}_j^{B_l}=\{\bar{y}_{ij}^{B_l}\}_{i=1}^n$ for each sentence $\mathbf{x}^{\mathcal{T}}_j\in D_{\mathcal{T}}$, respectively. Self-training is formulated as training the student network on the set $D_{\mathcal{S}} \cup D_d \cup D_a$, where the sets $D_d$ and $D_a$ are defined as follows:

\begin{align}
\begin{split}
    &D_d := \{(\mathbf{x}_j^{\mathcal{T}}, \bar{\mathbf{y}}_j^{B_l} )| \exists x_{ij} \in \mathbf{x}_j^{\mathcal{T}}\,\, {\rm s.t. }\,\, \bar{y}_{ij}^{{B_l}} \neq  \bar{y}_{ij}^{{A_l}}\} \\
    &D_a := \{(\mathbf{x}_j^{\mathcal{T}}, \bar{\mathbf{y}}_j^{B_l})| \forall  x_{ij} \in \mathbf{x}_j^{\mathcal{T}}\,\, {\rm s.t. }\,\, \bar{y}_{ij}^{{B_l}} =  \bar{y}_{ij}^{{A_l}}\}\\
\end{split}
\end{align}
Here, $\bar{y}_{ij}^{A_l}\in \bar{\mathbf{y}}_j^{A_l}$ is the teacher network's pseudo-label assignment on $x_{ij}\in \mathbf{x}_j^{\mathcal{T}}$. Let $r$ index the self-training round. Then the self-training loss for the student network at a specific self-training round $r$ can be formulated as follows:

% The self-training loss for the student network can then be formulated as \textcolor{red}{(Here the use of ${\mathcal{L}}^{B}_y$ is a problem. It should be distinguished from the one in equ (4). At least, they are different on the number of feeding samples. One is in the first training stage and this one is in the self-training stage.)}:

{
\begin{align}
    \begin{split}
        \bar{\mathcal{L}}^{B(r)}_y = {\mathcal{L}}^{B(r)}_y + \frac{1}{|D_d^{(r)}|}\sum_{(\mathbf{x}_j^{\mathcal{T}}, \bar{\mathbf{y}}_j^{B_l})\in D_d^{(r)}}\sum_{x_{ij}\in\mathbf{x}_j^{\mathcal{T}}}\ell (P(\hat{y}^{B_l}_{ij}), \bar{y}_{ij}^{B_l}) \\
        + \eta \frac{1}{|D_a^{(r)}|}\sum_{(\mathbf{x}_j^{\mathcal{T}}, \bar{\mathbf{y}}_j^{B_l})\in D_a^{(r)}}\sum_{x_{ij}\in\mathbf{x}_j^{\mathcal{T}}}\ell (P(\hat{y}^{B_l}_{ij}), \bar{y}_{ij}^{B_l})
    \end{split}
\end{align}
}

\noindent where $r\geq 1$, $\eta \in [0,1]$ is a variable to control the weight of the loss on $D_a^{(r)}$. Since the similarity between source and target domains can only be measured empirically, $\eta$ is treated as a hyper-parameter to be tuned. $\eta$ is expected to be large when the source and target domains are similar, otherwise small. Notice that when $\eta=1$, $\bar{\mathcal{L}}^{B(r)}_y$ becomes a special case of the pseudo-labelling loss function expressed in Eq. 15 of \cite{lee2013pseudo} with $\alpha(t)=1$, which we refer to as a standard pseudo-labelling method.

The total loss function $\mathcal{L}$ for SCD can now be formulated as 

%\textcolor{red}{(add a sentence to clarify; it includes the basic training then the self-training)}:

\begin{equation}
\begin{aligned}
\mathcal{L}&=\mathcal{L}_{d}^{B} + \mathcal{L}^{B(0)}_{y} + \sum_{r\geq 1}\bar{\mathcal{L}}^{{B}(r)}_y
\end{aligned}
\label{eqn:loss-adv}
\end{equation}
 In each self-training round, $D^{(r)}_{pl}=D_d^{(r)}\cup D_a^{(r)}$ is generated using the current trained student network. The self-training stops when $D^{(r)}_{pl}$ is approximately equal in successive rounds.

\input{Tables/results_table}
\input{Tables/results_ablation}

\section{Experiments and Results}\label{experiment}
%In this section, we perform experiments to validate our model on the cross-domain OTE task.

\subsection{Experimental Setup}
\subsubsection{Comparison Methods} We evaluate SCD as well as our BERT-based version BERT-SCD in this section. Comparison methods include, CRF~\cite{DBLP:conf/emnlp/JakobG10}, FEMA  \cite{DBLP:conf/naacl/YangE15}, Hier-Joint~\cite{DBLP:conf/aaai/DingYJ17}, RNSCN~\cite{DBLP:conf/acl/PanW18}, AD-SAL~\cite{DBLP:conf/emnlp/LiLWBZY19}, AHF~\cite{zhou2021adaptive} as well as the BERT-based models BERT$_{\rm E}$-UDA~\cite{DBLP:conf/emnlp/GongYX20} and BERT$_{\rm E}$-CDRG~\cite{DBLP:conf/acl/YuGX21}. Two strong single-domain OTE models BERT$_{\rm B}$ and BERT$_{\rm E}$~\cite{DBLP:conf/emnlp/GongYX20}, which are trained only on the source-domain to investigate the capacity of BERT without domain adaptation. SemBridge~\cite{DBLP:conf/acl/0002Q20} is excluded in our comparison since its dataset setup is different from that used in compared works.

\subsubsection{Datasets} We use benchmark datasets from four domains following previous work~\cite{DBLP:conf/acl/PanW18,DBLP:conf/emnlp/LiLWBZY19}. The Laptop dataset consists of reviews in the laptop domain taken from the SemEval ABSA challenge 2014~\cite{DBLP:conf/semeval/PontikiGPPAM14}. The Restaurant dataset is the set of all restaurant reviews in SemEval ABSA challenge 2014, 2015 and 2016~\cite{DBLP:conf/semeval/PontikiGPPAM14,DBLP:conf/semeval/PontikiGPMA15,DBLP:conf/semeval/PontikiGPAMAAZQ16}. The Device dataset, originally provided by~\cite{DBLP:conf/kdd/HuL04} contains reviews in the device domain. The Service dataset, introduced by \cite{DBLP:conf/acl/ToprakJG10} contains reviews related to the web service domain. We use the preprocessed data provided by~\cite{DBLP:conf/emnlp/LiLWBZY19}. Dataset statistics are shown in Table~\ref{tab:data_stat}.

\begin{table}[h]
% \small
    \centering
\begin{tabular}{c|c|c|c|c}\hline
Dataset &Domain & Sentence & Train & Test \\   \hline
$\mathbb{L}$ & Laptop & 1869 & 1458 & 411 \\ 
$\mathbb{R}$ & Restaurant & 3900 & 2481  & 1419\\
$\mathbb{D}$ & Device &1437 &954 & 483\\
$\mathbb{S}$ & Service &2153 &1433 & 720\\
 \hline
\end{tabular}
    \caption{Statistics of the datasets.}
    \label{tab:data_stat}
\end{table}

\subsubsection{Evaluation Protocol}
We follow prior work~\cite{DBLP:conf/emnlp/LiLWBZY19,DBLP:conf/emnlp/GongYX20} and evaluate on 10 transfer pairs $D_\mathcal{S} \rightarrow D_\mathcal{T}$ from the datasets. We use the test set of the source domain as a development set to tune our models. The test set of the target domain is used for evaluation purposes. We evaluate an exact match,\footnote{Exact Match: the predicted label sequence should exactly match the gold label sequence} and compute the Micro-F1 score. Reported results are the average over $5$ runs.

\subsubsection{Implementation Details}Following \citet{zhou2021adaptive}, we use $100$-dim fixed pretrained Word2Vec emebeddings~\cite{mikolov2013distributed} or BERT-Mini embeddings for word features.\footnote{We use BERT-Mini implementation from \url{https://github.com/google-research/bert}} We use Adam with $1e^{-3}$ learning rate, 100 epochs for both Teacher and Student networks, and 50 epochs during self-training, word embedding dropout rate in $[0.3, 0.5, 0.7]$, BiLSTM dimensions in $[100, 200, 300]$, adaption rate $\lambda \in [1.0, 0.7, 0.5, 0.3]$, batch size in $[32, 64, 128]$ and $\eta \in [0.0, 0.1,\ldots, \\ 0.9, 1.0, 1e^{-2}, 1e^{-3}]$. Each batch contains half labeled source and half unlabelled target data. All sentences are padded to a max length $n$. During self-training, we adopt repeated sampling on the labeled source data with the same size as the pseudo labeled target data in each epoch.

%\subsection{Results}
\subsection{Main Results} Table~\ref{tab:results} summarizes our main results. We find that neural methods, including RNSCN and Hier-Joint surpass hand-crafted feature methods FEMA and CRF, highlighting the importance of leveraging neural networks for the task. We also find that adversarial methods such as AD-SAL and AHF outperforms both Hier-Joint and RNSCN, indicating that adversarial learning is effective in mitigating the domain shift to yield performance. However, by learning target discriminative features, the SOTA method AHF achieves a better performance over AD-SAL by about 5.45 F1 on average. We see similar performance on the SOTA BERT-based model BERT$_{\rm E}$-CDRG that consider learning target discriminative features . Specifically, BERT$_{\rm E}$-CDRG outperforms the previous SOTA BERT$_{\rm E}$-UDA by about 4.47 F1 on average. This clearly shows the importance of learning target discriminative features. However, AHF considers the a mean teacher while BERT$_{\rm E}$-CDRG considers a generation model to learn these target discriminative features. In contrast, we consider to learn an adversarial model (i.e., Student) based on self-training through classifier disagreement. Our results suggest the effectiveness of our approach where we outperform AHF and BERT$_{\rm E}$-CDRG by an average F1 of 5.92 and 7.08. In particular, we obtain SOTA results on nine out of 10 transfer pairs with relative stability when compared to AHF.

%\textcolor{red}{Thus, learning on few but highly reliable target samples.}

\subsection{Ablation Study} We study the contribution of model components. Table~\ref{tab:ablation_results} presents our results. The upper portion of the table shows the performance of different ablated models. The lower portion is the Maximum Mean Discrepancy (MMD) \cite{gretton2012kernel}, which measures the distance between source and target domain distributions.\footnote{MMD from https://github.com/easezyc/deep-transfer-learning/} 

First, we note that the Teacher and Student networks have comparable performance on the source domain (see results in Table~\ref{tab:source-results}). This means the performance of the Student over Teacher is due to the divergence (measured by MMD). Since Student(MMD) is lower than Teacher(MMD) for all transfer pairs, it is not surprising to see the Student network outperforming the Teacher network. Conversely, SCD($\eta=1.0$) is simply standard pseudo-labelling. Although it improves performance, we find that  SCD($\eta=0.0$) offers comparable performance for the average F1 by focusing on learning only on pseudo-labelled samples with prediction disagreement with the Teacher network.  Interestingly, we find that on pairs such as  $\mathbb{S}\rightarrow\mathbb{L}$ and $\mathbb{S}\rightarrow\mathbb{D}$, Teacher(MMD) is already low. Although Student(MMD) becomes smaller due to adversarial learning, SCD($\eta=0.0$) cannot leverage sufficient pseudo-labelled samples to achieve satisfactory performance. This is because Student can only shift few samples to the domain invariant-distribution to bring about a prediction disagreement. But we see the benefit of prediction disagreement on pairs such as $\mathbb{D}\rightarrow\mathbb{R}$, where Teacher(MMD) is large and corresponding Student(MMD) is low, improving the Student network from 56.52 to 61.85 (i.e., performance on SCD($\eta=0.0$)).  

These results indicate that the pseudo-labelled samples help to learn the discriminative features, achieving better performance as compared to recent works. 

%\textcolor{red}{By paying attention in the learning problem}, SCD achieves better performance as compared to other model variants. 

\subsection{Sensitivity of Hyperparameter $\eta$} We now study the sensitivity of our model for the hyperparameter $\eta$. At $\eta=0$, we pay attention to the learning of pseudo-labelled samples by the student network that disagree with those produced by the Teacher network. At $\eta=1$, we are simply performing the standard pseudo-labelling. We study the sensitivity of $\eta$, particularly on pairs that have a high or low MMD on the Teacher network. That is, the respective $\mathbb{D}\rightarrow\mathbb{R}$ and $\mathbb{S}\rightarrow\mathbb{D}$ pairs. With low MMD, the source and target domains are similar, but diverges with high MMD. The idea is to understand how the domain divergence affects $\eta$. 

\begin{figure}[h]
	\centering
	\renewcommand\tabcolsep{0pt}
	\begin{tabular}{c c}
	\input{D_R_eta.tex}&\input{S_D_eta.tex}
	\end{tabular}
	\caption{F1 Performance of SCD for different $\eta$ values on $\mathbb{D}\rightarrow\mathbb{R}$ (left) and $\mathbb{S}\rightarrow\mathbb{D}$ (right) .}
	\label{fig:beta_fig}
\end{figure}
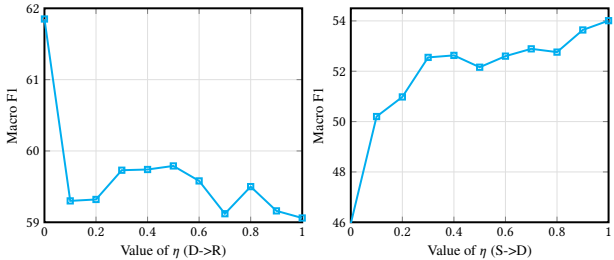

Figure \ref{fig:beta_fig} shows the results on this experiment where we report the F1 performance for different values of $\eta$ on the pairs. We find that on $\mathbb{S}\rightarrow\mathbb{D}$, the learning problem moves toward standard pseudo-labelling since the best performance is achieved at $\eta=1.0$. However, on $\mathbb{D}\rightarrow\mathbb{R}$ the best performance is achieved at $\eta=0$. These results suggest the importance of attention placed on the learning of these pseudo-labelled samples. Particularly, we observe that when the domain divergence is high it is beneficial to learn on pseudo-labelled samples that disagree with the Teacher network. On the other hand, when the source and target domains are similar, pseudo-labelling seems sufficient for the problem. This model behaviour guides in the selection of $\eta$.

\begin{figure*}[htbp]
	\centering
	\renewcommand\tabcolsep{10pt}
	\begin{tabular}{c c c}
	\scalebox{0.225}{\includegraphics[]{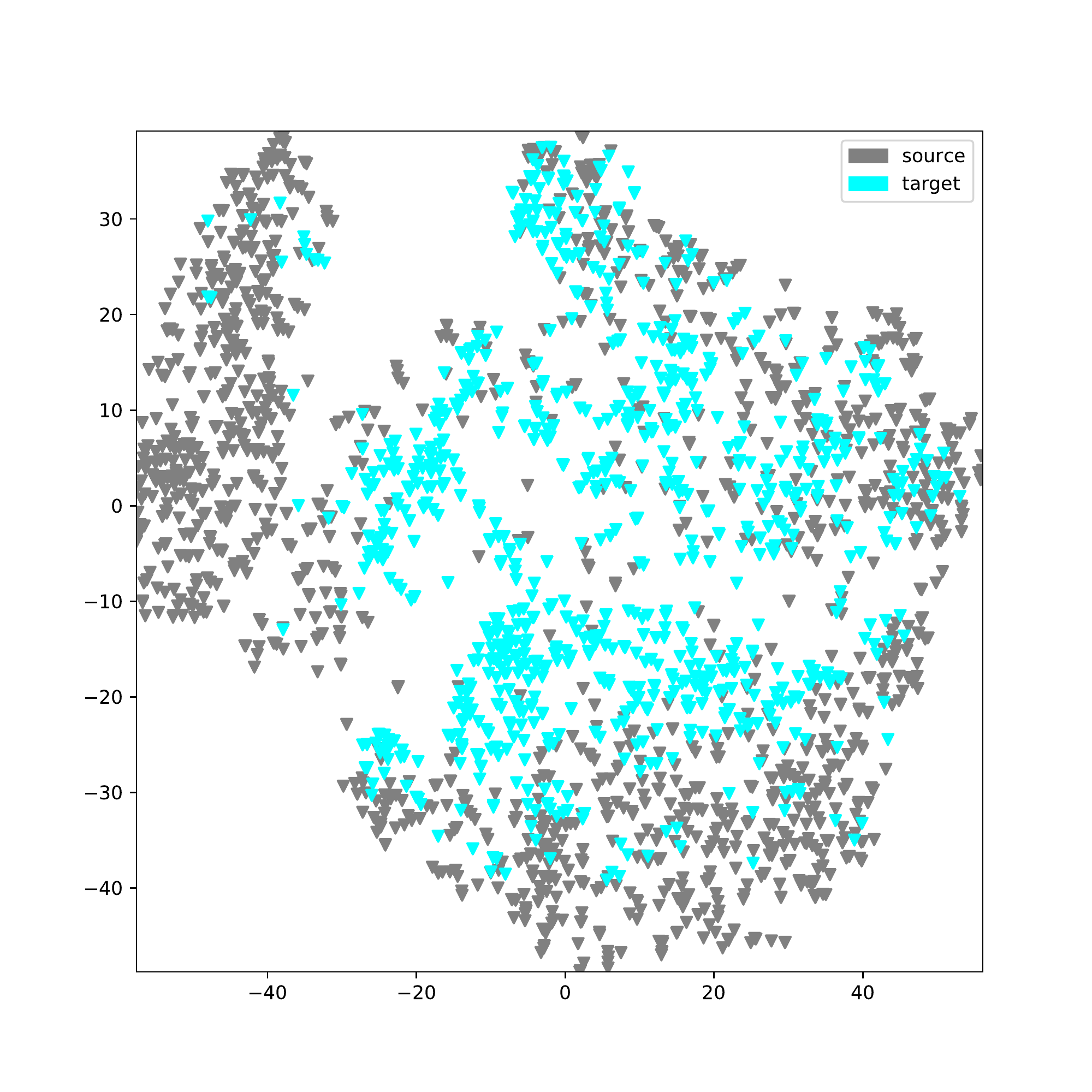}}
	&\scalebox{0.225}{\includegraphics[]{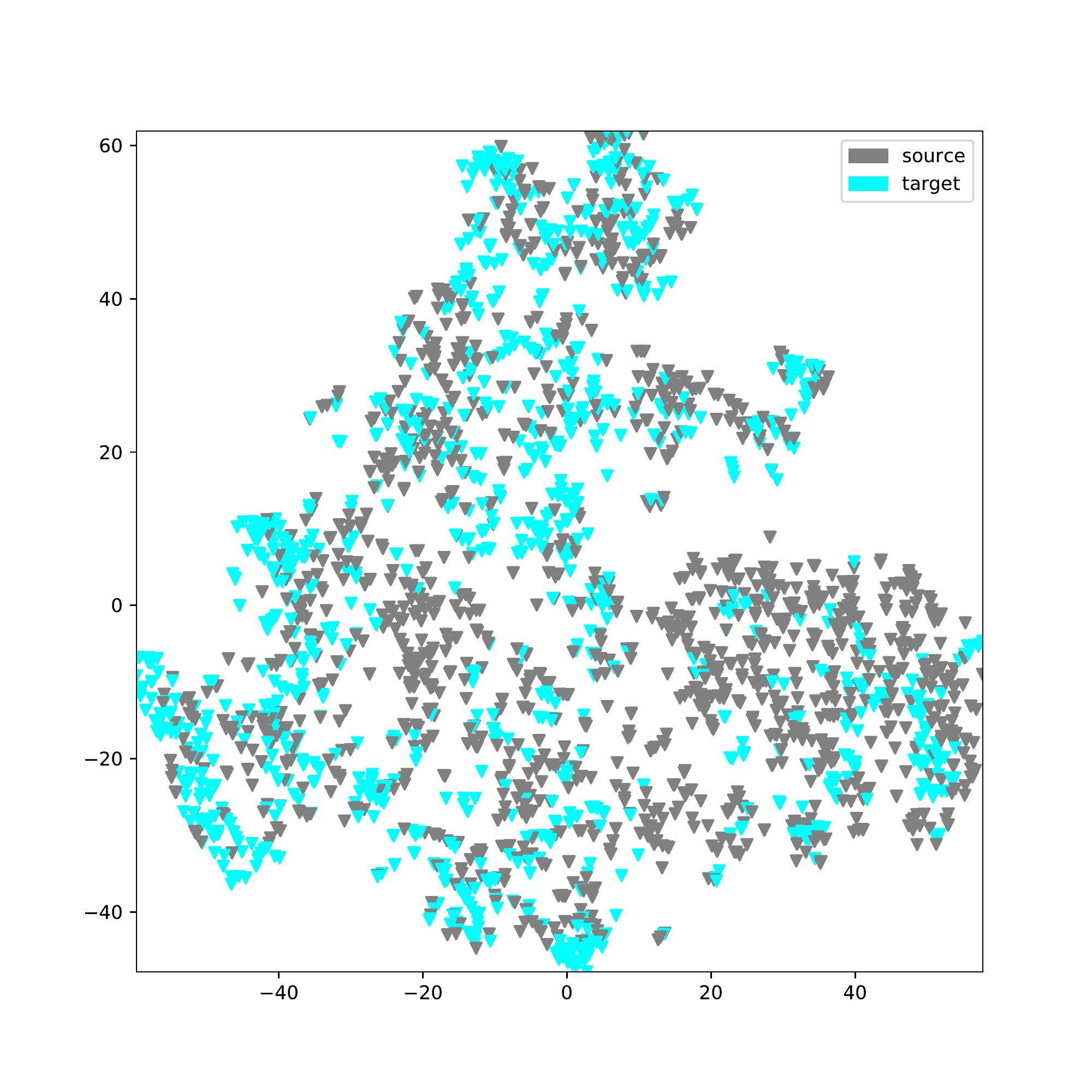}}
	&\scalebox{0.225}{\includegraphics[]{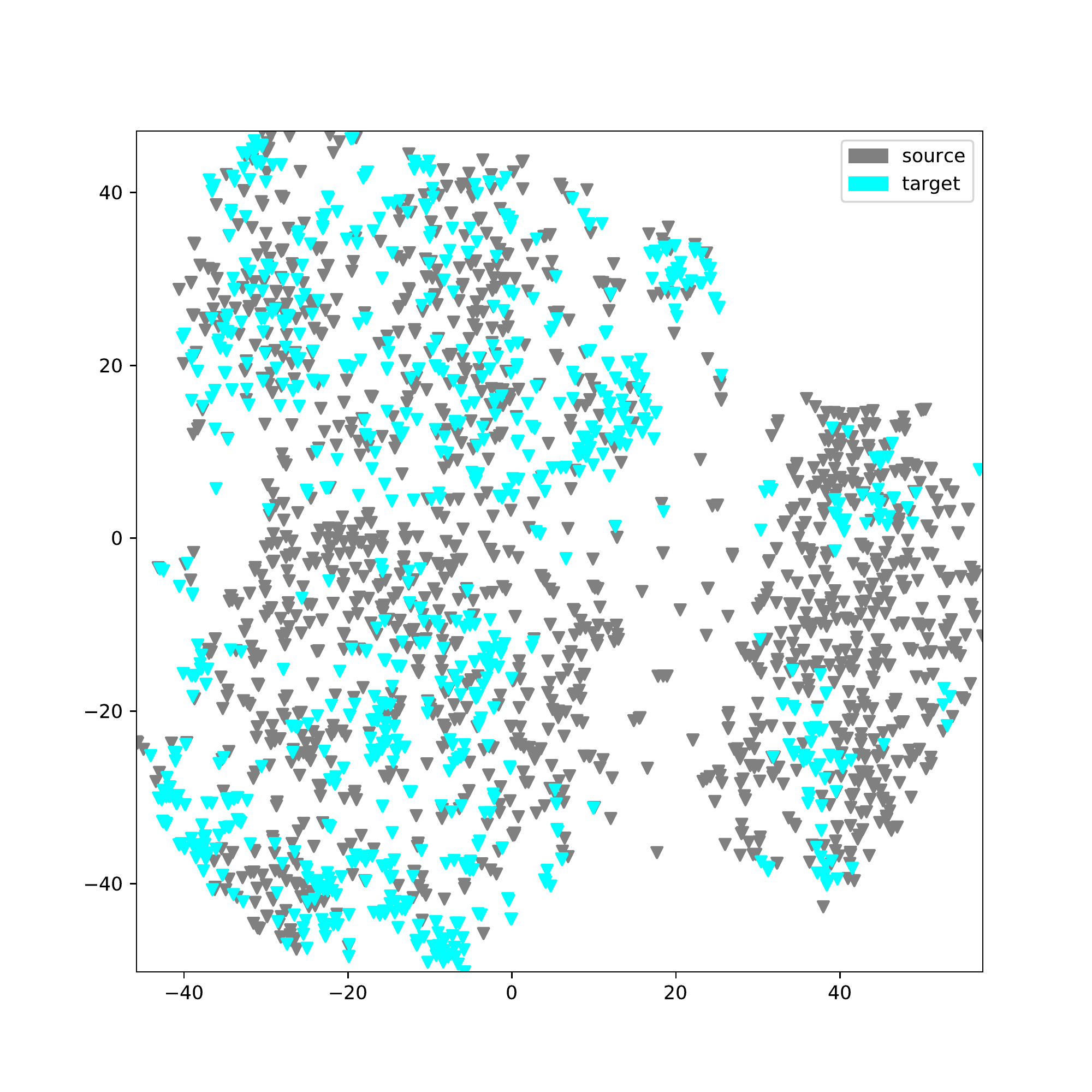}}\\
	(a) Teacher & (b) Student & (c) SCD \\
	\end{tabular}
	
	\caption{The t-SNE visualization of features learned by the (a) Teacher, (b) Student, and (c) SCD for the transfer pair $\mathbb{D}\rightarrow\mathbb{R}$ (light shade: target, dark shade: source) }
	\label{fig:vis}
\end{figure*}

% \textcolor{red}{(Though the model looks sensitive to the value of $\eta$, the MMD provides clues for model to select $\eta$. We believe our method can be easily employed in the real application.)}

\subsection{Quality of Pseudo-Labels}
\begin{figure}[h!]
	\centering
	\renewcommand\tabcolsep{0pt}
	\begin{tabular}{c c}
	\input{D_R.tex}&\input{S_D.tex}
	\end{tabular}
	\caption{F1 performance of different models for training epochs, aiming to evaluate the quality of pseudo labels.}
	\label{fig:quality_fig}
\end{figure}
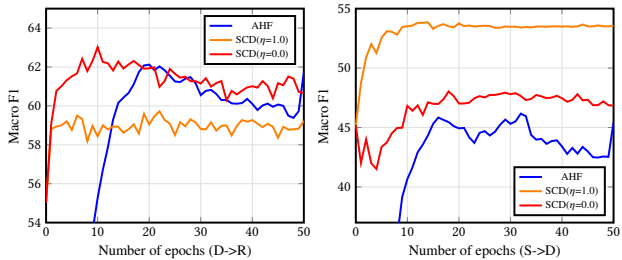

We perform additional experiments to study the quality of pseudo-labels generated by our method. Figure~\ref{fig:quality_fig} shows the experiments, where we report the F1 performance for different models for the pairs under study; $\mathbb{D}\rightarrow\mathbb{R}$ (left) and $\mathbb{S}\rightarrow\mathbb{D}$ (right). Since SCD($\eta=0.0$) and SCD have equivalent performance on $\mathbb{D}\rightarrow\mathbb{R}$ and SCD($\eta=1.0$) and SCD have equivalent performance on $\mathbb{S}\rightarrow\mathbb{D}$, we omit the curves of SCD to clearly show the benefit of pseudo-labelled samples under different strategies. Other compared methods include AHF. 

On $\mathbb{D}\rightarrow\mathbb{R}$, we find that both SCD($\eta=0.0$) and SCD($\eta=1.0$) improves steeply but becomes unstable after the fifth and eight epochs respectively. However, the improvement of SCD($\eta=0.0$) over SCD($\eta=1.0$) is highly notable. This observation points us to the fact, with high Teacher(MMD), prediction disagreement offers high quality pseudo-labelled samples particularly in the early rounds of training to improve performance. However, when Teacher(MMD) is low such as on the $\mathbb{S}\rightarrow\mathbb{D}$, we are not able to take advantage of pseudo-labelled samples with prediction disagreement. Hence, the standard pseudo-labelling can outperform prediction disagreement as seen in the figure. AHF on the other hand underperforms, indicating that our SSL approach is effective as compared to the mean teacher.

\subsection{Feature Visualization} Fig.~\ref{fig:vis} depicts the t-SNE~\cite{van2008visualizing} visualization of features learned using the Teacher, Student and SCD models on the transfer pair $\mathbb{D}\rightarrow\mathbb{R}$ (1000 instances sampled randomly in each domain). As there are three class labels, namely BIO labels, an ideal model should clearly align the source and target data into three clusters. For the Teacher network, we can observe that the distribution of source samples is relatively far from the distribution of the target samples. Through domain adaptation, the Student network improves the alignment of the source and target samples. However, by learning target discrimnative features through SCD, we gradually observe three clusters forming. The results indicate that SCD improves the class-level alignment.

\input{Tables/case-study}

\begin{table*}[htbp]
    % \scriptsize
    \centering
    \begin{tabular}{l|c|c|c|c|c|c|c|c|c|c|c} \hline
         Model &$\mathbb{S}\rightarrow\mathbb{R}$ &$\mathbb{L}\rightarrow\mathbb{R}$ &$\mathbb{D}\rightarrow\mathbb{R}$ &$\mathbb{R}\rightarrow\mathbb{S}$ &$\mathbb{L}\rightarrow\mathbb{S}$ &$\mathbb{D}\rightarrow\mathbb{S}$ &$\mathbb{R}\rightarrow\mathbb{L}$ &$\mathbb{S}\rightarrow\mathbb{L}$ &$\mathbb{R}\rightarrow\mathbb{D}$& $\mathbb{S}\rightarrow\mathbb{D}$&AVG \\ \hline
      
           Teacher &69.63&76.35&66.73 &82.67&75.43&67.59 &80.00&69.20 &79.74&69.16 &73.65$\pm$0.60 \\
           
           Student &67.48&77.88&65.72 &82.63&74.44&67.53 &80.17&70.03 &80.64&69.41 &73.59$\pm$0.62 \\
          
        \hline
    \end{tabular}
    \caption{F1 performance of Teacher and Student on the test set of the source domain.}
    \label{tab:source-results}
\end{table*}
% \section{Appendix}
% \label{sec:appendix}

\subsection{Case Study}

To test the effectiveness of our approach, some case examples from the transfer pair with the largest domain divergence ($\mathbb{D}\rightarrow\mathbb{R}$) are selected for demonstration. Table~\ref{tab:case-study} shows the aspect term extraction results on these case examples.

% We perform a case study on SCD, the Teacher and Student networks to test the effectiveness of our approach. Sentences are selected from the $\mathbb{D}\rightarrow\mathbb{R}$ transfer pair, which has the largest domain divergence between the Device and Restaurant domains (see the score for Teacher(MMD) in Table~\ref{tab:ablation_results}). Table~\ref{tab:case-study} shows the opinion target extraction results on our case examples.

In the first case, we find that the Teacher, Student and SCD are all capable of identifying the aspect terms ``service'' and ``space''. As these aspect terms appear in both Device and Restaurant domains,  domain adaptation is not necessary to extract the aspect terms. It is therefore not surprising to observe that all models identify the aspect terms in the Restaurant domain. 

In the second case example, the aspect terms ``ambience'', ``food'' and ``catfish'' are found in the Restaurant domain and not the Device domain. However, the Teacher was able to extract the aspect terms ``ambience'' and ``food''. Introspecting further, we found that 81\% of aspect terms extracted by the Teacher  in the Restaurant domain are accompanied with opinion words (e.g., ``great'') that are also present in the Device domain. Hence, the Teacher was able to learn the correspondences between opinion words and aspect terms in the Device domain and use that knowledge to locate ``ambience'' and ``food'' in the Restaurant domain. However, both Teacher and Student networks fail to extract the aspect term ``catfish''. This highlights the importance of learning target discriminative features, as there is no correspondence between the word ``delicious'' and an aspect term to be learned in the Device domain but only in the Restaurant domain. SCD solves this problem by collecting high quality pseudo-labelled samples in the Restaurant domain. As a result, SCD is able to extract the aspect term ``catfish''.

In the third case example, we found that the Teacher network failed to identify the aspect terms ``pasta primavera'' and ``veggies'' as they do not exist in the Device domain. However, by reducing the domain shift between the two domains, the Student network is able to extract ``pasta primavera'' but not ``veggies''. Upon investigation, we found that the opinion word ``fresh'' which expresses an opinion on ``veggies'' frequently appears 83 times in the Restaurant dataset and 0 times in the Device dataset. Ideally, by learning target discriminative features, we can learn correspondences that exist between ``fresh'' and aspect terms. Such knowledge as learned by SCD offers supervisory training signals, enabling SCD to detect the aspect term ``veggies''. 

%This means that the Student network may require to learn target discriminative features to improve its performance.

Finally, in the fourth case example, both the Teacher and Student networks completely failed to detect the aspect term ``martinis''. While it is no surprise that the Teacher network fails (i.e., ``martinis'' is not seen during training), the failure of the Student network highlights the limitations of simply reducing the domain shift and suggests the importance of  learning target discriminative features for successful cross-domain OTE.

\subsection{Performance Comparison on Source Domain}
We argued that the difference between the target errors (or F1 performance) of the teacher and student networks can be explained by the $\mathcal{H}\Delta\mathcal{H}$ divergence when the source errors of these networks are approximately equal. According to \citet{ben2010theory}, the source error as well as the divergence can be estimated from finite samples of the source and target domains, under the assumption of the uniform convergence theory \cite{vapnik2015uniform}. Table~\ref{tab:source-results} therefore reports the F1 performance on the source test set. We discover that for each transfer pair, the F1 performance is approximately equal, comparing the Teacher and Student. This suggest that adversarial learning performed by the Student to reduce the domain shift has little to no effect on the classification on the source data. Most importantly, the results suggest that the difference between the Teacher and Student on the target data is due to the target samples shifted to the domain-invariant space within the Student feature space.

\section{Conclusion}

We have proposed a Self-training through Classifier Disagreement for cross-domain OTE. We demonstrated that by simultaneously training a Teacher and a Student network, we can benefit from the information that comes from their predictions on the unlabelled target domain. Specifically, by leveraging pseudo-labelled samples that disagree between the Teacher and Student networks, the Student network is significantly improved, even in large domain divergences. This model behaviour however leads to the potential limitation. In cases of small domain shifts, the model tends to favor pseudo-labeling~\cite{lee2013pseudo}, an SSL approach that risks confirmation bias \cite{tarvainen2017mean} (i.e., prediction errors are fit by the network). Nevertheless, small domain shifts have little to no interest in cross-domain learning since the source and target domains can be considered to be similar. In the future, we will consider data augmentation strategies to mitigate confirmation bias brought by pseudo-labelling in such situations \cite{arazo2020pseudo}. We believe our model is generic and can be applied to other cross-domain tasks such as cross-domain named entity recognition. 

%That is, in small domain shifts, the model tends to lean towards performing pseudo-labelling~\cite{lee2013pseudo}; an SSL approach that risks confirmation bias \cite{tarvainen2017mean} (i.e., prediction errors are fit by the network).

%We found that by attending to learning on pseudo-labelled samples by the Student network that disagree with those made by the Teacher network, we can significantly improve the Student network. This benefits comes from a large domain divergence and the relatedness among the source and target domains. 

\section*{Acknowledgements}
This work was supported in part by the National Key R\&D Program of China under Grant 2021ZD0110700, in part by the Fundamental Research Funds for the Central Universities, in part by the State Key Laboratory of Software Development Environment. In addition, SM and NA received support from the Leverhulme Trust under Grant Number: RPG\#2020\#148.

\bibliographystyle{ACM-Reference-Format}
\bibliography{sample-base}

\end{document}

%% file: tikz_package.tex
\usepackage{tikz}
\usetikzlibrary{decorations.pathreplacing}
\usetikzlibrary{fadings}
\usetikzlibrary{arrows}
\usepackage{rotating}
\usetikzlibrary{fit}
\usepackage{pgfplots}

\usetikzlibrary{%
  arrows,%
  shapes,
  shapes.misc,% wg. rounded rectangle
  shapes.arrows,%
  chains,%
  matrix,%
  positioning,% wg. " of "
  scopes,%
  decorations.pathmorphing,% /pgf/decoration/random steps | erste Graphik
  shadows,
}

\tikzset{
  nonterminal/.style={
    rectangle,
    minimum size=6mm,
    very thick,
    draw=red!50!black!50,         % 50% red and 50% black,
    top color=white,              % a shading that is white at the top...
    bottom color=red!50!black!20, % and something else at the bottom
    font=\itshape
  },
  terminal/.style={
    rounded rectangle,
    minimum size=6mm,
    very thick,draw=black!50,
    top color=white,bottom color=black!20,
    font=\ttfamily},
  skip loop/.style={to path={-- ++(0,#1) -| (\tikztotarget)}}
}

{
  \tikzset{terminal/.append style={text height=1.5ex,text depth=.25ex}}
  \tikzset{nonterminal/.append style={text height=1.5ex,text depth=.25ex}}
}

\pgfplotsset{compat=1.5}

%% file: model_pseudo.tex
    \begin{tikzpicture}[-latex ,auto ,node distance =2 cm and 2 cm ,on grid ,
semithick , 
state/.style ={ circle ,top color =white , bottom color = processblue!20 ,
draw, processblue , text=black , minimum width =0.025cm, text width=0.025cm},
box/.style ={rectangle ,top color =white , bottom color = processblue!20 ,
draw, processblue , text=blue , minimum width =0.5cm , minimum height = 0.5cm},
dbox/.style ={rectangle ,top color =white , bottom color = processblue!20 ,
draw, processblue , text=blue , minimum width =0.5cm , minimum height = 0.5cm},
neuron/.style ={rectangle ,top color =white , bottom color = red!20 ,
draw, red , text=red , minimum width =0.5cm , minimum height = 0.5cm, rounded corners},
triangle/.style = {top color =white , bottom color = processblue!20 ,
draw, processblue , text=blue, regular polygon, regular polygon sides=3, minimum size=0.5cm, draw },
node rotated/.style = {rotate=270},
    border rotated/.style = {shape border rotate=270},
gn/.style={trapezium, trapezium angle=67.5, draw, inner ysep=5pt, outer sep=0pt, minimum height=1.81mm, minimum width=0pt}
    ]

\definecolor{target}{RGB}{255,190,6}
    \definecolor{source}{RGB}{0,178,236}

    \begin{scope}[shift={(0.1,-0.15)}]
        \node[fill=source, circle, thick, draw=black!80,inner sep=0pt,minimum size=7pt] at (1.5, 3.5)  () {};
        \node[fill=source, circle, thick, draw=black!80,inner sep=0pt,minimum size=7pt] at (1, 3.5)  () {};
        \node[fill=source, circle, thick, draw=black!80,inner sep=0pt,minimum size=7pt] at (2, 3.5)  () {};
        
        \node[fill=source, regular polygon,regular polygon sides=3, thick, draw=black!80,inner sep=0pt,minimum size=7pt] at (1.5, 3.)  () {};
        \node[fill=source, regular polygon,regular polygon sides=3, thick, draw=black!80,inner sep=0pt,minimum size=7pt] at (1, 3.)  () {};
        \node[fill=source, regular polygon,regular polygon sides=3, thick, draw=black!80,inner sep=0pt,minimum size=7pt] at (2, 3.)  () {};
    \end{scope}
    \draw [-, line width=0.8pt, rounded corners=0mm](-1.1,3.75)--(-1.1,2.15)--(2.5,2.15)--(2.5,3.75)--(-1.1,3.75);
    \draw [-, dashed, line width=0.8pt, rounded corners=0mm](0.8,3.6)--(0.8,2.65)--(2.4,2.65)--(2.4,3.6)--(0.8,3.6);
    \draw [-, dashed, line width=0.8pt, rounded corners=0mm](-1.0,3.6)--(-1.0,2.65)--(0.6,2.65)--(0.6,3.6)--(-1.0,3.6);
    \draw [-, dashed, line width=0.8pt, rounded corners=0mm](-0.45,2.65)--(-0.45,3.6);

    \begin{scope}[shift={(3.0,0)}]
        \node[fill=target, rectangle, thick, draw=black!80,inner sep=0pt,minimum size=7pt] at (2.5, 3.5)  () {};
        \node[fill=target,rectangle, thick, draw=black!80,inner sep=0pt,minimum size=7pt] at (2, 3.5)  () {};
        \node[fill=target,rectangle, thick, draw=black!80,inner sep=0pt,minimum size=7pt] at (3, 3.5)  () {};
        
        \node[fill=target,rectangle, thick, draw=black!80,inner sep=0pt,minimum size=7pt] at (2.5, 3.)  () {};
        \node[fill=target,rectangle, thick, draw=black!80,inner sep=0pt,minimum size=7pt] at (2, 3.)  () {};
        \node[fill=target,rectangle, thick, draw=black!80,inner sep=0pt,minimum size=7pt] at (3, 3.)  () {};
    \end{scope}
    
    \draw [-, dashed, line width=0.8pt, rounded corners=0mm](4.75,3.75)--(4.75,2.75)--(6.25,2.75)--(6.25,3.75)--(4.75,3.75);
    
     \node [draw=white] (Dt) at (5.5, 2.5){$D_\mathcal{T}$};
    
    \node [box, minimum height=0.65cm, minimum width=2.8cm](bilstm) at (1.5,4.75){Student Encoder $B_e$};
    \draw [->,line width=0.8pt, rounded corners=1mm, color=red](1.5, 3.75)--(1.5,4.4);
    \draw [->,line width=0.8pt, rounded corners=1mm, color=blue](5.5, 3.75)--(5.5,4.4);
    \draw [->,line width=0.8pt, rounded corners=0mm, color=blue](5.5, 4)--(2.25,4)--(2.25,4.4);
    
    \node [box,minimum width=2.6cm, minimum height=0.65cm](classifier) at (1.5,5.8){Student Classifier $B_l$};
    \node [](argmax) at (3.1,6.6){};
    \node [](argmax) at (6.1,6.6){};
    \draw [->,line width=0.8pt, rounded corners=1mm, color=red](1.5, 5.1)--(1.5, 5.45);
    \draw [->,line width=0.8pt, rounded corners=1mm, color=blue](2.25,5.1)--(2.25,5.45);

    \node [draw=white] (L) at (1.5, 6.9){$\bar{\mathcal{L}}^{B}_y$}; 
    \draw [->, line width=0.8pt, rounded corners=1mm, color=red](1.5, 6.15)--(L);
    
    \begin{scope}[shift={(0,6.5)}]
        \node[fill=target, regular polygon, regular polygon sides=3,  thick, draw=black!80,inner sep=0pt,minimum size=7pt] at (2.5, 1)  () {};
        \node[fill=target, circle, thick, draw=black!80,inner sep=0pt,minimum size=7pt] at (2., 1)  () {};
        \node[fill=target, regular polygon, regular polygon sides=3, thick, draw=black!80,inner sep=0pt,minimum size=7pt] at (3, 1)  () {};
        
        \node[fill=target, circle,  thick, draw=black!80,inner sep=0pt,minimum size=7pt] at (2.5, 1.5)  () {};
        \node[fill=target, regular polygon, regular polygon sides=3, thick, draw=black!80,inner sep=0pt,minimum size=7pt] at (2., 1.5)  () {};
        \node[fill=target, circle, thick, draw=black!80,inner sep=0pt,minimum size=7pt] at (3, 1.5)  () {};
    \end{scope}
    
    \draw [->,line width=0.8pt, rounded corners=1mm, color=blue](2.25,6.15)--(2.25,7.25);
    
    \draw [-, dashed, line width=0.8pt, rounded corners=0mm](1.75,8.25)--(1.75,7.25)--(3.25,7.25)--(3.25,8.25)--(1.75,8.25);
    
    \draw [-, dashed, line width=0.8pt, rounded corners=0mm](2.25,8.25)--(2.25,7.25);

    \node [box, minimum height=0.65cm, minimum width=2.8cm](bilstm2) at (5.5,4.75){Teacher Encoder $A_e$};
    \node [box, minimum height=0.65cm, minimum width=2.6cm](classifier2) at (5.5,5.8){Teacher Classifier $A_l$};
    \draw [->,line width=0.8pt, rounded corners=1mm, color=blue](bilstm2)--(classifier2);

    \begin{scope}[shift={(3.0,6.5)}]
        \node[fill=target, regular polygon, regular polygon sides=3, thick, draw=black!80,inner sep=0pt,minimum size=7pt] at (2.5, 1)  () {};
        \node[fill=target, regular polygon, regular polygon sides=3, thick, draw=black!80,inner sep=0pt,minimum size=7pt] at (2., 1)  () {};
        \node[fill=target, regular polygon, regular polygon sides=3, thick, draw=black!80,inner sep=0pt,minimum size=7pt] at (3, 1)  () {};
        
        \node[fill=target, circle, thick, draw=black!80,inner sep=0pt,minimum size=7pt] at (2.5, 1.5)  () {};
        \node[fill=target, circle, thick, draw=black!80,inner sep=0pt,minimum size=7pt] at (2., 1.5)  () {};
        \node[fill=target, circle, thick, draw=black!80,inner sep=0pt,minimum size=7pt] at (3, 1.5)  () {};
        
    \end{scope}
    
    \draw [->,line width=0.8pt, rounded corners=1mm, color=blue](5.5,6.15)--(5.5,7.25);
    
    \draw [-, dashed, line width=0.8pt, rounded corners=0mm](4.75,8.25)--(4.75,7.25)--(6.25,7.25)--(6.25,8.25)--(4.75,8.25);
    
    \draw [->, line width=0.8pt, rounded corners=1mm](4.6,7.75)--(3.35,7.75);
    \node [] (null1) at (4, 8){};
    \draw [<-, line width=0.8pt, rounded corners=1mm](4.6,7.6)--(3.35,7.6);
    \node [] (null1) at (4, 8){};
    
    \draw [->, line width=0.8pt, rounded corners=0.1mm](1.75,7.75)--(-0.45,7.75)--(-0.45,3.6);
    
     \node [] (DD) at (1.9, 8.5){$D_d$};
     \node [] (DA) at (2.72, 8.5){$D_a$};

    \usetikzlibrary{shapes.geometric}
    
    \begin{scope}[shift={(6.25,5)}]
        \node[fill=source, regular polygon,regular polygon sides=3, thick, draw=black!80,inner sep=0pt,minimum size=5pt] at (1.35, 3.5)  () {};
        \node[fill=source, circle, thick, draw=black!80,inner sep=0pt,minimum size=5pt] at (1, 3.5)  () {};
        \node [] (source) at (2.65, 3.5){\small{Labeled Source}};
    \end{scope}
    
    \begin{scope}[shift={(6.25,4.65)}]
        \node[fill=target, regular polygon, regular polygon sides=3, thick, draw=black!80,inner sep=0pt,minimum size=5pt] at (1.35, 3.5)  () {};
        \node[fill=target, circle, thick, draw=black!80,inner sep=0pt,minimum size=5pt] at (1, 3.5)  () {};
        \node [] (target) at (2.63, 3.5){\small{Labeled Target}};
    \end{scope}

    \begin{scope}[shift={(6.25,4.3)}]
        \node[fill=target, rectangle, thick, draw=black!80,inner sep=0pt,minimum size=5pt] at (1.2, 3.5)  () {};
        
        \node [draw=white] (unlabel) at (2.65, 3.5){\small{Unlabel Target}};
    \end{scope}
    
    \begin{scope}[shift={(6.25,3.95)}]
        \node[fill=white, circle, thick, draw=black!80,inner sep=0pt,minimum size=5pt] at (1.2, 3.5)  () {};
     
        \node [draw=white] (classA) at (2.2, 3.5){\small{Class 1}};
    \end{scope}
    
    \begin{scope}[shift={(6.25,3.6)}]
        \node[fill=white, regular polygon,regular polygon sides=3, thick, draw=black!80,inner sep=0pt,minimum size=5pt] at (1.2, 3.5)  () {};

        \node [draw=white] (classB) at (2.2, 3.5){\small{Class 2}};
    \end{scope}

    \draw [-, line width=0.8pt, rounded corners=0mm](7.,8.75)--(7.,6.85)--(10.,6.85)--(10.,8.75)--(7.,8.75);

     \begin{scope}[shift={(-2.7,1.85)}]
        \node[fill=target, regular polygon,regular polygon sides=3,  thick, draw=black!80,inner sep=0pt,minimum size=7pt] at (2.5, 1)  () {};
        \node[fill=target, circle, thick, draw=black!80,inner sep=0pt,minimum size=7pt] at (2., 1)  () {};
        \node[fill=target, regular polygon,regular polygon sides=3, thick, draw=black!80,inner sep=0pt,minimum size=7pt] at (3, 1)  () {};
        
        \node[fill=target, circle,  thick, draw=black!80,inner sep=0pt,minimum size=7pt] at (2.5, 1.5)  () {};
        \node[fill=target, regular polygon,regular polygon sides=3, thick, draw=black!80,inner sep=0pt,minimum size=7pt] at (2., 1.5)  () {};
        \node[fill=target, circle, thick, draw=black!80,inner sep=0pt,minimum size=7pt] at (3, 1.5)  () {};
    \end{scope}

    \node [] (Dd) at (-0.7, 2.4){$D_d$};
    \node [] (Da) at (0.1, 2.4){$D_a$};
    \node [] (Ds) at (1.6, 2.4){$D_\mathcal{S}$};
    
     \draw [->, line width=0.8pt, rounded corners=0mm, color=red](7.5,6.5)--(8.5,6.5);
     \node [] (Train) at (9, 6.5){\textcolor{red}{Train}};
     
      \draw [->, line width=0.8pt, rounded corners=0mm, color=blue](7.5,6)--(8.5,6);
     \node [] (Predict) at (9.1, 6){\textcolor{blue}{Predict}};
     
    \draw [->, line width=0.8pt, rounded corners=1mm](7.5,5.5)--(8.5,5.5);
    \node [] (null1) at (9.3, 5.4){Compare};
    \draw [<-, line width=0.8pt, rounded corners=1mm](7.5,5.35)--(8.5,5.35);

\end{tikzpicture}

%% file: Tables/results_table.tex
\begin{table*}[htbp]
    % \scriptsize
    \centering
    \begin{tabular}{l|ccc|ccc|cc|cc|c} \hline
         Model &$\mathbb{S}\rightarrow\mathbb{R}$ &$\mathbb{L}\rightarrow\mathbb{R}$ &$\mathbb{D}\rightarrow\mathbb{R}$ &$\mathbb{R}\rightarrow\mathbb{S}$ &$\mathbb{L}\rightarrow\mathbb{S}$ &$\mathbb{D}\rightarrow\mathbb{S}$ &$\mathbb{R}\rightarrow\mathbb{L}$ &$\mathbb{S}\rightarrow\mathbb{L}$
         &$\mathbb{R}\rightarrow\mathbb{D}$
         &$\mathbb{S}\rightarrow\mathbb{D}$
         &AVG\\ \hline
         CRF &17.00&17.00&2.50&8.80&8.60&4.50&10.90&11.60&9.00&9.70 &9.96 \\
         FEMA               &37.60&35.00&20.70&10.80&14.80&8.80&26.60&15.00&22.90&18.70 &21.09\\
         Hier-Joint         &52.00&46.70&50.40&19.80&23.40&23.50&31.70&30.00&32.00&33.40 &34.29 \\
         RNSCN              &48.89&52.19&50.39&30.41&31.21&35.50&47.23&34.03&46.16&32.41 &40.84 \\
         AD-SAL             &52.05&56.12&51.55&39.02&38.26&36.11&45.05&35.99&43.76&41.21 &43.91 \\
        
        AHF &54.98&58.67&61.11 &40.33&47.17&45.78 &56.58&36.62 &48.24&44.16 &49.36$\pm$3.23\\
         \hline

        SCD &\bf59.52&\bf71.40&\bf61.85 &\bf48.30&\bf48.67&\bf52.58 &\bf59.68&\bf42.40 &\bf54.45&\bf54.01 &\bf55.28$\pm$1.07 \\
         \hline \hline 
         
         BERT$_{\rm B}$ &54.29&46.74&44.63 &22.31&30.66&33.33 &37.02&36.88 &32.03&38.06 &37.60\\
         BERT$_{\rm E}$ &57.56&50.42&45.71 &26.50&25.96&30.40 &44.18&41.78 &35.98&35.13 &39.36\\
         BERT$_{\rm E}$-UDA &59.07&55.24&56.40&34.21&30.68&38.25& 54.00& 44.25&42.40&40.83 &45.53\\
         
         BERT$_{\rm E}$-CDRG &59.17&\bf68.62&58.85&47.61&\bf54.29&42.20&55.56 &41.77&35.43&36.53 &50.00\\
        
       \hline 
       BERT-SCD &\bf64.10&67.61&\bf64.75 &\bf55.83&51.33&\bf58.92 &\bf55.64&\bf49.76 &\bf49.62&\bf53.29 &\bf57.08$\pm$1.17\\ \hline
    \end{tabular}
    \caption{Comparison of F1 performance. Best performance is in bold format.}
    \label{tab:results}
\end{table*}

%% file: Tables/results_ablation.tex
\begin{table*}[htbp]
    % \scriptsize
    \centering
    \begin{tabular}{l|ccc|ccc|cc|cc|c} \hline
         Model &$\mathbb{S}\rightarrow\mathbb{R}$ &$\mathbb{L}\rightarrow\mathbb{R}$ &$\mathbb{D}\rightarrow\mathbb{R}$ &$\mathbb{R}\rightarrow\mathbb{S}$ &$\mathbb{L}\rightarrow\mathbb{S}$ &$\mathbb{D}\rightarrow\mathbb{S}$ &$\mathbb{R}\rightarrow\mathbb{L}$ &$\mathbb{S}\rightarrow\mathbb{L}$
         &$\mathbb{R}\rightarrow\mathbb{D}$
         &$\mathbb{S}\rightarrow\mathbb{D}$
         &AVG
          \\
         \hline
        
         SCD &\bf 59.52&\bf71.40&\bf 61.85 &\bf 48.30&\bf 48.67&\bf 52.58 &\bf59.68&\bf 42.40 &\bf54.45&\bf54.01 &\bf 55.28$\pm$1.07 \\
         SCD($\eta=0.0$) &59.18&\bf 71.40&\bf 61.85 &48.22&48.52&52.25 &57.81&40.13  &52.78&45.95 &53.80$\pm$1.91 \\
         SCD($\eta=1.0$) &57.76&67.49&59.06 &47.83&46.13&51.03 &55.62&\bf 42.40 &53.80&\bf 54.01 &53.51$\pm$0.96 \\

         Student &55.39&63.69&56.52 &47.19&45.48&50.69 &52.66&41.22 &52.39&44.28 &50.95$\pm$1.23 \\
         Teacher &52.10&57.46&48.02 &24.88&28.48&33.09 &48.08&40.92 &50.75&45.35 &42.87$\pm$1.10 \\
         \hline
          \hline
           Student(MMD) &0.041&0.040&0.046 &0.035&0.094&0.080 &0.054&0.042 &0.045&0.043 &0.052$\pm$0.009\\
           Teacher(MMD) &0.215&0.197&0.415 &0.364&0.170&0.263 &0.198&0.134 &0.158&0.106 &0.222$\pm$0.023\\
          
         \hline 
    \end{tabular}
    \caption{{{\bf Ablation Study:} F1 Performance of different ablated models (top). Student(MMD) (or Teacher(MMD)) is an estimate of the discrepancy between the learned source and target distributions by the Student (or Teacher).}}
    \label{tab:ablation_results}
\end{table*}

%% file: D_R_eta.tex
\begin{tikzpicture}[scale=0.5]
\begin{axis}[
 legend style={font=\small, legend pos=north east},
 label style={font=\Large},
 tick label style={font=\Large},
	xmin=0,xmax=1,
	ymin=59,ymax=62,
	xlabel=Value of $\eta$ (D->R), % \hertz requires SIunits
	ylabel=Macro F1,
	grid=both,
	minor grid style={gray!25},
	major grid style={gray!25},
	line width=1.5pt,
	]
\addplot[ color=cyan, mark=square] coordinates {
(0.0,61.85)
(0.1,59.3)
(0.2,59.32)
(0.3,59.73)
(0.4,59.74)
(0.5,59.79)
(0.6,59.58)
(0.7,59.12)
(0.8,59.5)
(0.9,59.16)
(1.0,59.06)
};
\end{axis}
\end{tikzpicture}

%% file: S_D_eta.tex
\begin{tikzpicture}[scale=0.5]
\begin{axis}[
 legend style={font=\small, legend pos=north east},
 label style={font=\Large},
 tick label style={font=\Large},
	xmin=0,xmax=1,
	ymin=46,ymax=54.5,
	xlabel=Value of $\eta$ (S->D), % \hertz requires SIunits
	ylabel=Macro F1,
	grid=both,
	minor grid style={gray!25},
	major grid style={gray!25},
	line width=1.5pt,
	]
\addplot[ color=cyan, mark=square] coordinates {
(0.0,45.95)
(0.1,50.2)
(0.2,50.98)
(0.3,52.55)
(0.4,52.63)
(0.5,52.16)
(0.6,52.6)
(0.7,52.89)
(0.8,52.76)
(0.9,53.64)
(1.0,54.01)
};
\end{axis}
\end{tikzpicture}

%% file: D_R.tex
\begin{tikzpicture}[scale=0.5]
\begin{axis}[
 legend style={font=\small, legend pos=north east},
 label style={font=\Large},
 tick label style={font=\Large},
	xmin=0,xmax=50,
	ymin=54.,ymax=65,
	xlabel=Number of epochs (D->R), % \hertz requires SIunits
	ylabel=Macro F1,
	grid=both,
	minor grid style={gray!25},
	major grid style={gray!25},
	line width=1.5pt,
	]
\addplot[ color=blue] coordinates {
(0,0.0)
(1,0.0)
(2,0.0)
(3,0.0)
(4,0.0)
(5,1.0421189751202578)
(6,26.237989650011595)
(7,42.145593865794616)
(8,50.861313864177816)
(9,53.53535353075668)
(10,55.267423009865404)
(11,56.70157067582411)
(12,57.88667687109815)
(13,59.29648240717243)
(14,60.15297310142424)
(15,60.42636608181701)
(16,60.66282420254174)
(17,61.10979928664989)
(18,61.73933317291933)
(19,62.07852193496994)
(20,62.120525595369216)
(21,61.835315229466204)
(22,62.03090507226321)
(23,61.83156526886924)
(24,61.551687078603344)
(25,61.256995259745004)
(26,61.225373452458754)
(27,61.39554978977275)
(28,61.47876697064222)
(29,61.162341835018886)
(30,60.55243849305862)
(31,60.77614557710923)
(32,60.82106202088351)
(33,60.61705988611435)
(34,60.29919447142379)
(35,60.305697077000254)
(36,60.13011151918354)
(37,60.112359545587125)
(38,60.14375144412595)
(39,60.358291221471426)
(40,60.10528724609841)
(41,59.78582820188992)
(42,60.02290950245728)
(43,60.11534024876403)
(44,59.95348836711363)
(45,60.070011663634496)
(46,60.00470255792562)
(47,59.491565687601565)
(48,59.386152743076494)
(49,59.72915181257188)
(50,61.73933317291933)
};
\addplot[ color=orange] coordinates {
 (0,55.04)
(1,58.809999999999995)
(2,58.940000000000005)
(3,59.0)
(4,59.209999999999994)
(5,58.76)
(6,59.5)
(7,59.330000000000005)
(8,58.19)
(9,58.97)
(10,58.440000000000005)
(11,59.0)
(12,58.79)
(13,58.95)
(14,58.919999999999995)
(15,58.620000000000005)
(16,58.8)
(17,59.050000000000004)
(18,58.58)
(19,59.589999999999996)
(20,59.050000000000004)
(21,59.46)
(22,59.72)
(23,59.27)
(24,59.089999999999996)
(25,58.5)
(26,59.160000000000004)
(27,58.95)
(28,59.309999999999995)
(29,59.150000000000006)
(30,58.809999999999995)
(31,58.79)
(32,59.160000000000004)
(33,58.730000000000004)
(34,58.97)
(35,58.989999999999995)
(36,58.48)
(37,58.97)
(38,59.27)
(39,59.19)
(40,59.27)
(41,59.089999999999996)
(42,58.91)
(43,59.08)
(44,58.93000000000001)
(45,58.36)
(46,58.93000000000001)
(47,58.78)
(48,58.79)
(49,58.830000000000005)
(50,59.230000000000004)
};

\addplot[ color=red] coordinates {
(0,55.04)
(1,59.089999999999996)
(2,60.77)
(3,61.019999999999996)
(4,61.31999999999999)
(5,61.519999999999996)
(6,61.67)
(7,62.419999999999995)
(8,61.79)
(9,62.3)
(10,63.019999999999996)
(11,62.260000000000005)
(12,62.19)
(13,61.82)
(14,62.28)
(15,61.919999999999995)
(16,62.129999999999995)
(17,62.31)
(18,62.1)
(19,61.91)
(20,61.92999999999999)
(21,61.980000000000004)
(22,60.980000000000004)
(23,61.28)
(24,61.9)
(25,61.650000000000006)
(26,61.45)
(27,61.5)
(28,61.27)
(29,61.27)
(30,61.150000000000006)
(31,61.419999999999995)
(32,60.99)
(33,61.17)
(34,61.28)
(35,60.29)
(36,60.77)
(37,60.589999999999996)
(38,60.78)
(39,60.95)
(40,61.01)
(41,60.940000000000005)
(42,61.27)
(43,61.019999999999996)
(44,60.39)
(45,61.17)
(46,61.06)
(47,61.51)
(48,61.4)
(49,60.72)
(50,60.650000000000006)
};

 \addlegendentry{\small AHF}
 \addlegendentry{\small SCD($\eta$=1.0)}
 \addlegendentry{\small SCD($\eta$=0.0)}
\end{axis}
\end{tikzpicture}

%% file: S_D.tex
\begin{tikzpicture}[scale=0.5]
\begin{axis}[
 legend style={font=\small, legend pos=south east},
 label style={font=\Large},
 tick label style={font=\Large},
	xmin=0,xmax=50,
	ymin=37.,ymax=55,
	xlabel=Number of epochs (S->D), % \hertz requires SIunits
	ylabel=Macro F1,
	grid=both,
	minor grid style={gray!25},
	major grid style={gray!25},
	line width=1.5pt,
	]
\addplot[ color=blue] coordinates {
(0,0.0)
(1,0.0)
(2,0.0)
(3,0.0)
(4,3.0219780211626386)
(5,25.645933009911747)
(6,30.744336564660983)
(7,32.08556149234729)
(8,35.69277107934969)
(9,39.15756629769026)
(10,40.659340654385026)
(11,41.653418119064796)
(12,42.91369754058032)
(13,43.56120826214905)
(14,44.33887568784872)
(15,45.310015893309625)
(16,45.836637584270534)
(17,45.64183834687721)
(18,45.45454544960777)
(19,45.11517076551023)
(20,44.92979718692584)
(21,44.9648711894183)
(22,44.16796266999638)
(23,43.706563701592934)
(24,44.54685099348887)
(25,44.69696969198541)
(26,44.32757324820428)
(27,44.64419475156407)
(28,45.23809523310216)
(29,45.66345440568014)
(30,45.307917883565466)
(31,45.52129221233022)
(32,46.165191735416904)
(33,45.94992635730101)
(34,44.378257627174)
(35,43.97905758663212)
(36,44.042232272539536)
(37,43.622641504447515)
(38,43.85832704851689)
(39,43.92097263939447)
(40,43.40490797048394)
(41,42.78074865812268)
(42,43.284727546827924)
(43,42.80215549926076)
(44,43.3716475046018)
(45,42.98780487306831)
(46,42.498095958461526)
(47,42.47517188195767)
(48,42.56292905680493)
(49,42.53048779989758)
(50,45.45454544960777)
};
\addplot[ color=orange] coordinates {
 (0,45.190000000000005)
(1,48.72)
(2,50.9)
(3,52.019999999999996)
(4,51.27)
(5,52.54)
(6,53.1)
(7,53.05)
(8,52.81)
(9,53.449999999999996)
(10,53.52)
(11,53.580000000000005)
(12,53.800000000000004)
(13,53.81)
(14,53.849999999999994)
(15,53.300000000000004)
(16,53.55)
(17,53.690000000000005)
(18,53.510000000000005)
(19,53.43)
(20,53.76)
(21,53.53)
(22,53.580000000000005)
(23,53.49)
(24,53.54)
(25,53.559999999999995)
(26,53.54)
(27,53.39)
(28,53.39)
(29,53.5)
(30,53.47)
(31,53.49)
(32,53.410000000000004)
(33,53.449999999999996)
(34,53.43)
(35,53.49)
(36,53.5)
(37,53.47)
(38,53.510000000000005)
(39,53.510000000000005)
(40,53.49)
(41,53.510000000000005)
(42,53.47)
(43,53.510000000000005)
(44,53.66)
(45,53.510000000000005)
(46,53.510000000000005)
(47,53.59)
(48,53.510000000000005)
(49,53.510000000000005)
(50,53.55)
};

\addplot[ color=red] coordinates {
 (0,45.190000000000005)
(1,42.02)
(2,44.0)
(3,41.99)
(4,41.510000000000005)
(5,43.37)
(6,43.730000000000004)
(7,44.519999999999996)
(8,44.96)
(9,44.98)
(10,46.82)
(11,46.410000000000004)
(12,46.86)
(13,46.06)
(14,47.12)
(15,46.989999999999995)
(16,46.98)
(17,47.370000000000005)
(18,48.03)
(19,47.64)
(20,47.02)
(21,47.02)
(22,47.089999999999996)
(23,47.46)
(24,47.64)
(25,47.53)
(26,47.74)
(27,47.74)
(28,47.83)
(29,47.93)
(30,47.8)
(31,47.89)
(32,47.65)
(33,47.31)
(34,47.4)
(35,47.74)
(36,47.620000000000005)
(37,47.47)
(38,47.52)
(39,47.67)
(40,47.46)
(41,47.17)
(42,47.32)
(43,47.8)
(44,47.29)
(45,47.33)
(46,46.88)
(47,46.88)
(48,47.199999999999996)
(49,46.88)
(50,46.83)
};

 \addlegendentry{\small AHF}
 \addlegendentry{\small SCD($\eta$=1.0)}
 \addlegendentry{\small SCD($\eta$=0.0)}
\end{axis}
\end{tikzpicture}

%% file: Tables/case-study.tex
% \begin{table*}[htbp]
%     % \footnotesize
%     \centering
%     \begin{tabular}{p{0.5cm}|p{5cm}|p{2.5cm}|p{2.5cm}|p{2.5cm}}
%     \hline
%         Case &Sentence & Teacher & Student & SCD \\
%         \hline
%          1 & For the amount of {\bf food} we got, the {\bf prices} should have been lower. & [prices] & [food, prices] & [food, prices]\\ \hline
%          2 & The {\bf service} was excellent, the {\bf coffee} was good even by starbucks standards and the {\bf food} was outstanding. &[service]&[service, food]&[service, coffee, food]\\ \hline
%          3 & The {\bf music} is great, no night better or worse, the {\bf bar tenders} are generous with the pouring, and the lighthearted {\bf atmosphere} will lifts you spirits. &[music, atmosphere]&[music, atmosphere]&[music, bar tenders, atmosphere]\\ \hline
%     \end{tabular}
%     \caption{Case study on the transfer pairs $\mathbb{L}\rightarrow\mathbb{R}$. Gold opinion targets are boldfaced.}
%     \label{tab:case-study}
% \end{table*}

\begin{table*}[htbp]
    % \footnotesize
    % \centering
    \begin{tabular}{p{0.5cm}|p{5.5cm}|p{2.1cm}|p{2.1cm}|p{3.2cm}}
    \hline
        Case &Sentence & Teacher & Student & SCD \\
        \hline
         1 & But the \textbf{space} is small and lovely, and the \textbf{service} is helpful. &space, service&space, service&space, service\\ \hline
         2 & Although small, it has beautiful \textbf{ambience}, excellent \textbf{food} and \textbf{catfish} is delicious. &ambience, food&ambience, food&ambience, food, catfish\\ \hline
         % 3 & We had a very hard time getting the \textbf{waitress} attention and finally had to get up and go inside to speak to a manager.  &[] &[waitress] &[waitress]\\ \hline
         3 &The \textbf{pasta primavera} was outstanding as well, lots of fresh \textbf{veggies} &NULL &pasta primavera &pasta primavera, veggies\\ \hline
         4 & I would definitely go back, if only for some of those exotic \textbf{martinis} on the blackboard. &NULL &NULL &martinis\\ \hline
    \end{tabular}
    \caption{Case study on $\mathbb{D}\rightarrow\mathbb{R}$. Gold aspect terms are boldfaced. ``NULL'' indicates that no aspect term has been extracted.}
    \label{tab:case-study}
\end{table*}